\documentclass[conference]{IEEEtran}
\IEEEoverridecommandlockouts
\usepackage{amsmath,amssymb,amsfonts}
\usepackage{algorithmic}
\usepackage{graphicx}
\usepackage{textcomp}
\usepackage{cite}
\usepackage{siunitx}
\usepackage{xcolor}
\usepackage[capitalize]{cleveref}
\usepackage{tabulary}
\usepackage{booktabs}
\usepackage[acronym]{glossaries}
\def\BibTeX{{\rm B\kern-.05em{\sc i\kern-.025em b}\kern-.08em
    T\kern-.1667em\lower.7ex\hbox{E}\kern-.125emX}}
\glsdisablehyper

\newcommand{\imgscale}{1}
\newcommand{\rnum}[1]{\num[round-mode=figures,round-precision=3]{#1}}
\newcommand{\numfeatures}{356}
\newcommand{\xgbcSpecAtMinSeventySens}{66.32653061224489}
\newcommand{\svcSpecAtMinSeventySens}{48.97959183673469}
\newcommand{\lrcSpecmAtMinSeventySens}{42.85714285714286}
\newcommand{\xgbcThresAtMinSeventySens}{0.2563808}
\newcommand{\percentpositivetvt}{38.77038895859473}
\newcommand{\calibrationoddsratio}{0.8230414606196789}

\newacronym{svc}{SVM}{support vector machine}
\newacronym{roc}{ROC}{receiver operating characteristic}
\newacronym{auc}{AUC}{area under curve}
\newacronym{mcc}{MCC}{Matthews correlation coefficient}
\newacronym{upm}{UPM}{unified performance measure}
\newacronym{lowess}{LOWESS}{locally weighted scatterplot smoothing}
\newacronym{bmi}{BMI}{body mass index}
\newacronym{nuss}{NUSS}{Non-Union Scoring System}
\newacronym{rust}{RUST}{Radiographic Union Scale of Tibial Fractures}
\newacronym{ml}{ML}{machine learning}
\newacronym{oddsratio}{OR}{odds ratio}

\usepackage{fancyhdr}
\fancypagestyle{firstpage}{
    \fancyhead[L]{46th Annual International Conference of the IEEE Engineering in Medicine \& Biology Society (EMBC 2024)}
    \fancyfoot[C]{\footnotesize © 2024 Authors. DOI: \href{https://doi.org/10.1109/EMBC53108.2024.10782650}{10.1109/EMBC53108.2024.10782650}}
}
\begin{document}

\title{Predictive Model Development to Identify Failed Healing in Patients after Non–Union Fracture Surgery%
\thanks{This work has received funding from the European Research Council (ERC) Consolidator Grant Safe data-driven control for human-centric systems (CO-MAN) under Grant Agreement No. 864686.}
}

\author{%
\IEEEauthorblockN{Cedric Donié\textsuperscript{*}, Marie K. Reumann\textsuperscript{\dag}, Tony Hartung\textsuperscript{\dag}, Benedikt J. Braun\textsuperscript{\dag}, Tina Histing\textsuperscript{\dag}, Satoshi Endo\textsuperscript{*},\\Sandra Hirche\textsuperscript{*}}
\IEEEauthorblockA{Email: \{cedric.donie,s.endo,hirche\}@tum.de\{mreumann,thartung,bbraun,thisting\}@bgu-tuebingen.de}
\IEEEauthorblockA{\textsuperscript{*}Chair of Information-Oriented Control, Technical University of Munich, Munich, Germany}
\IEEEauthorblockA{\textsuperscript{\dag}Dept. of Trauma and Reconst. Surg., BG Klinik Tuebingen, Eberhard Karls University Tuebingen, Tuebingen, Germany}
}

\maketitle
\thispagestyle{firstpage}

\begin{abstract}
Bone non-union is among the most severe complications associated with trauma surgery, occurring in 10--\SI{30}{\percent} of cases after long bone fractures. Treating non-unions requires a high level of surgical expertise and often involves multiple revision surgeries, sometimes even leading to amputation. Thus, more accurate prognosis is crucial for patient well-being.

Recent advances in machine learning (ML) hold promise for developing models to predict non-union healing, even when working with smaller datasets, a commonly encountered challenge in clinical domains. To demonstrate the effectiveness of ML in identifying candidates at risk of failed non-union healing, we applied three ML models---logistic regression, support vector machine, and XGBoost---to the clinical dataset TRUFFLE, which includes 797 patients with long bone non-union.

The models provided prediction results with 70\% sensitivity, and the specificities of
\SI[round-mode=figures,round-precision=2]{\xgbcSpecAtMinSeventySens}{\percent} (XGBoost),
\SI[round-mode=figures,round-precision=2]{\svcSpecAtMinSeventySens}{\percent} (support vector machine),
and \SI[round-mode=figures,round-precision=2]{\lrcSpecmAtMinSeventySens}{\percent} (logistic regression).
These findings offer valuable clinical insights because they enable early identification of patients at risk of failed non-union healing after the initial surgical revision treatment protocol.
\end{abstract}


\section{Introduction}
\begin{figure*}
    \centering
    \includegraphics[width=6.21in]{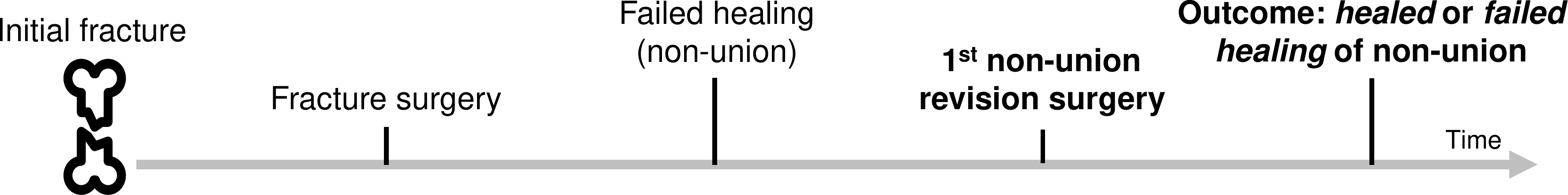} %
    \caption{Timeline of the clinical treatment of long bone non-unions}
    \label{fig:timeline}
\end{figure*}

Bone non-union describes the failed healing of a fracture and represents one of the most severe complications encountered in the field of trauma surgery. The subsequent treatment necessitates complex surgical interventions, leading to a substantial reduction in patient quality of life~\cite{brinkerDevastatingEffectsHumeral2022}. 
Non-union treatment is challenging and frequently requires revision surgeries (\cref{fig:timeline}).
In some cases, revisions fail, making amputation the only option. Consequently, non-unions are a major socioeconomic burden~\cite{saulBoneHealingGone2023}. Such treatment failures are partially attributable to the limited understanding of the factors that influence bone healing after non-union treatment. Predicting the outcome of the first non-union revision surgery would, therefore, be helpful to design more personalized interventions and increase the chance of recovery.
Developing such a prediction model is challenging because the collection of non-union data is time-consuming and expensive, limiting the size of datasets.

The clinical need to predict whether patients will (continue to) suffer from a non-union is currently handled by the surgical expertise of individual specialists. Various heuristics have been developed to formalize this knowledge for the use in clinical practice. However, heuristic prediction methods have limited power due to their inability to capture complex patterns.

A novel approach is the use of \gls{ml} to predict failed healing in patients with non-union after bone fracture.
\Gls{ml} works well for risk and outcome prediction in other areas of medicine~\cite{elfanagelyMachineLearningSurgical2021}, and it has been used in prospective studies to predict whether non-union would occur after fractures. ML can predict the healing of vertebral fractures based on features such as medical imaging findings, \gls{bmi}, and age~\cite{takahashiMachinelearningbasedApproachNonunion2022,leisterPredictiveModelIdentify2023} and for subtrochanteric femoral fractures based on age and treatment type~\cite{nagPreclinicalModelPostsurgery2022}. However, this research only predicts non-union after very specific fractures. To our knowledge, no study has investigated predicting the healing of existing non-unions for different anatomical locations in long bones.

In this paper, we propose a model that successfully predicts failed healing after the first revision surgery for non-unions in various long bones.
Our approach is investigated and validated with a dataset that was collected in a single trauma hospital.

\section{Methods}
\subsection{Patient cohort}
We collected data from 797 patients (\SI{67.9}{\percent} male, \SI{32.1}{\percent} female sex, self-reported) with long bone non-union fractures at a level I trauma center from 1\textsuperscript{st} January 2009 to 31\textsuperscript{st} May 2023 (TRUFFLE database, ClinicalTrials.gov NCT06098157).
Age ranged from 14 to 91 years (mean \num{50.1}, SD \num{15.6}).
We received ethical approval from the University of Tuebingen ethical committee (840/2019BO2).
A long bone non-union was defined as a fracture that had not shown any clinical and radiological signs of healing, with criteria adapted from the \gls{rust}~\cite{whelanDevelopmentRadiographicUnion2010}. %

\cref{fig:timeline} illustrates the timeline of patients undergoing non-union treatment. During the follow-up visit after the first non-union revision surgery, orthopaedic and trauma specialists evaluated the treatment outcome: non-union healed or failed to heal. In the present dataset, failed healing was observed in \SI[round-mode=places,round-precision=2]{\percentpositivetvt}{\percent} of patients.

The dataset contained clinical features related to the initial trauma (i.e. date, anatomical location of fracture, severity of injury), to primary surgical treatment (i.e. date, type of surgical reduction and fixation), to screening at the time non-union diagnosis (i.e. age, sex, comorbidities, body mass index, Weber-Cech classification, biomechanical stability, soft tissue status, lab results, drug intake), and to non-union revision surgery (i.e. date, type of fixation, autologous bone grafting, use of growth factors, need of antibiotic treatment).

\subsection{Data preprocessing}

Classification requires features to be numerical for many approaches (e.g., in scikit-learn). We encode nominal values (categorical, e.g., type of osteosynthesis) with one-hot encoding, taking into account that more than one choice is possible for some features (e.g., comorbidities of diabetes and lung disease). This produces our dataset containing \numfeatures{} features per patient with \SI{16.6}{\percent} missing data. All dates (e.g., of the first fracture surgery or the first non-union revision surgery) are counted from the initial fracture date.

We split the data into \SI{80}{\percent} training and \SI{20}{\percent} test data.
The proportion of patients with failed healing is equivalent for training (\SI{38.77}{\percent}) and test (\SI{38.75}{\percent}) data.
To avoid overfitting, we hold out the test data, reserving \SI{20}{\percent} of training data for intermediate model evaluation instead.

\subsection{Model development}
XGBoost, \glspl{svc}, and logistic regression were analyzed in the present work. Each of these models outputs a predicted probability of failed non-union healing. To convert the probability to a binary prediction (not healed vs. healed), a threshold was chosen and all probabilities above this threshold are were rated as failed healing. Unless stated otherwise, we used a threshold of 0.5.

XGBoost is an ensemble of \emph{regression trees}~\cite{chenXGBoostScalableTree2016}. Regression trees are similar to decision trees, but output an estimated likelihood rather than a binary prediction. XGBoost trains many regression trees on subsets of the training data via ~\emph{boosting} and averages their output to yield a class likelihood. Boosting means that new trees in the ensemble depend on previously trained trees.
We selected XGBoost for further study because it is state-of-the-art for tabular data classification~\cite{borisovDeepNeuralNetworks2022,chenXGBoostScalableTree2016}, can model non-linear relationships well~\cite{elfanagelyMachineLearningSurgical2021}, and was used in related non-union work~\cite{leisterPredictiveModelIdentify2023,takahashiMachinelearningbasedApproachNonunion2022}.

\Glspl{svc} classify high-dimensional input into two classes by treating each feature as a dimension and linearly separating the features with a \emph{hyperplane}. For example, two features would be separated by a line (1d hyperplane), three features by a plane (2d hyperplane), and $n$ features by a ($n-1$)-dimensional hyperplane~\cite{cortesSupportvectorNetworks1995}. Non-linear classification is possible by transforming the feature space with a radial base function. Probabilities are obtained by applying a logistic function to the \gls{svc}'s output~\cite{plattProbabilitiesSVMachines2023}. \Gls{svc} was chosen because it is common in contemporary research~\cite{elfanagelyMachineLearningSurgical2021} and it has seen time-testing than XGBoost.

Logistic regression outputs the probability of the positive class via the logistic function applied to a linear combination of all the features~\cite{steyerbergClinicalPredictionModels2019}. An optimization algorithm selects regression coefficients that minimize the prediction error. We chose logistic regression as a baseline because it is simple, the most common medical prediction model, and works comparatively well for the oftentimes small clinical datasets encountered in clinical settings~\cite{steyerbergClinicalPredictionModels2019}.

For XGBoost, the positive class is weighted with the ratio of healed to non-healed patients in the training data, compensating for the slight class imbalance. To reduce the chance of overfitting that would reduce the performance on external validation, we opted for the standard XGBoost hyperparameters, which are known to work on a wide variety of datasets~\cite{borisovDeepNeuralNetworks2022}.
However, we limited the tree depth to five, as performance deteriorated with deeper trees in our initial experiments.

Logistic regression and \gls{svc} benefit from scaled input and cannot handle missing values. Thus, standard scaling is applied to all ordinal (e.g., Weber-Czech~\cite{weberPseudarthrosisPathophysiologyBiomechanics1976} classification into hypertrophic, oligotrophic, and atrophic), interval (e.g., number of previous surgeries), and continuous (e.g., hemoglobin) values.
Standard scaling $z$ of a single value $x$ is defined as $z = (x - \mu)/\sigma$, where $\mu$ is the mean and $\sigma$ is the standard deviation over all patients.
Missing numerical values are replaced with the feature's mean and missing booleans/categoricals with the feature's most frequent value.

\subsection{Model evaluation}
For the evaluation of a classifier, the first step is to calculate a single metric summarizing the classifier's performance.
The next step is to compared the performance metrics between classifiers and consider statistical significance.
In addition to the discriminative ability of the model, it is critical to evaluate model calibration to understand how well the predicted risk of non-union matches the actual incidence of non-union. %

\subsubsection{Performance metrics}
The \gls{upm}~\cite{redondoUnifiedPerformanceMeasure2020} and the \gls{mcc}~\cite{chiccoMatthewsCorrelationCoefficient2023} are most frequently recommended~\cite{chiccoMatthewsCorrelationCoefficient2023}. We opted for the \gls{upm}, which can be calculated directly from the confusion matrix according to%
~\cite{redondoUnifiedPerformanceMeasure2020}.
\begin{align}
    \text{UPM} &= \frac{4 \cdot T P \cdot T N}{4 \cdot T P \cdot T N+(T P+T N) \cdot(\text{FP} + \text{FN})} \label{eq:upmconfusionmatrix} \nonumber \\ 
    &= \frac{4}{\frac{1}{\text { Precision }}+\frac{1}{\text { Sens. }}+\frac{1}{\text {Spec.}} + \frac{1}{\text{Neg. Predictive Value}}} \nonumber
\end{align}

We examined multiple thresholds to compensate for the fact that \gls{upm} depends on a single confusion matrix (i.e., threshold).

\subsubsection{Statistics for model comparison}
It is important to ensure that differences between several classifiers' \gls{upm} scores are statistically significant, requiring multiple trainings and evaluations of the same classifier. However, all classifiers are deterministic during training, which means that repeated training of the same model would yield identical scores. To provide a random component, we generated new training datasets by randomly sampling \SI{80}{\percent} of the full training dataset without replacement. This was repeated 300 times to generate 300 unique training datasets. We trained XGBoost, logistic regression, and \gls{svc} for each of these 300 datasets. The Wilcoxon signed-rank test compared the performance of the classifiers pairwise. The effect size $r$ was estimated as $r=Z/\sqrt{N}$, where $Z$ is the normalized test statistic and $N$ is the number of samples. We set the significance level at $\alpha=0.05$ and used Bonferroni correction for multiple comparisons.

\subsubsection{Model calibration}
To measure how well calibrated a model is, we regressed each patient's actual healing/failed healing against the predicted probability of failed healing by means of \gls{lowess} (with the \textit{statsmodels} implementation defaults of 2/3 of the total data for each $y$ and three residual-based re-weightings), as suggested by \cite{steyerbergClinicalPredictionModels2019}. The regression was applied to all of the test data. Furthermore, the mean bias of our model was expressed as an \gls{oddsratio} of predicted incidence $\hat{y}$ vs. actual incidence $y$.%
\begin{align}
\mathrm{OR} & = \operatorname{odds}(\operatorname{mean}(\hat{y})) / \operatorname{odds}\left(\operatorname{mean}\left(y\right)\right) \nonumber \\
& =\frac{\operatorname{mean}(\hat{y}) /\left(1-\operatorname{mean}(\hat{y})\right)}{\operatorname{mean}\left(y\right) /\left(1-\operatorname{mean}\left(y\right)\right)}
\nonumber
\end{align}

\subsection{Ablation studies}
To understand how the results would change with the amount of patient data, we artificially reduced the training data.
Specifically, we defined multiple fractions of the training data from zero to one.
For each of these fractions, we generated 25 samples (with replacement) from the entire training dataset, trained XGBoost, and evaluated the performance (\gls{upm}, sens., spec.) over the whole test dataset.

\section{Results and Discussion}\label{sec:results}
All three models delivered a good predictive performance with \SI{70}{\percent} sensitivity and specificities of
\SI[round-mode=figures,round-precision=2]{\xgbcSpecAtMinSeventySens}{\percent} for XGBoost,
\SI[round-mode=figures,round-precision=2]{\svcSpecAtMinSeventySens}{\percent} for \gls{svc},
and \SI[round-mode=figures,round-precision=2]{\lrcSpecmAtMinSeventySens}{\percent} for logistic regression.

\begin{figure}
    \centering
    \includegraphics[scale=\imgscale]{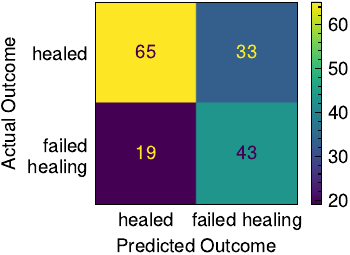}
    \caption{Confusion matrix for XGBoost with a threshold of \num[round-mode=figures,round-precision=2]{\xgbcThresAtMinSeventySens}. This threshold was chosen to guarantee at least \SI{70}{\percent} sensitivity.}
    \label{fig:xgboost_cm}
\end{figure}

\begin{figure}
    \centering
    \includegraphics[scale=\imgscale]{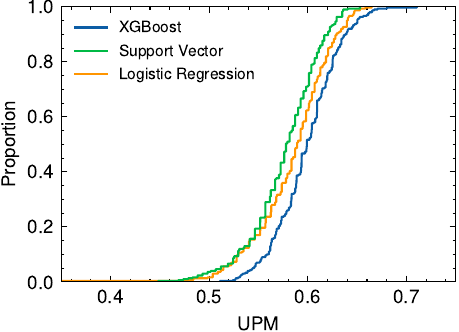}
    \caption{Empirical cumulative distribution functions of XGBoost, SVM, and logistic regression. Each classifier is trained 300 times on randomly sampled 80\% of the training data. A lower curve indicates stochastic dominance.}
    \label{fig:modelcomparison}
\end{figure}

\begin{figure}
    \centering
    \includegraphics[scale=\imgscale]{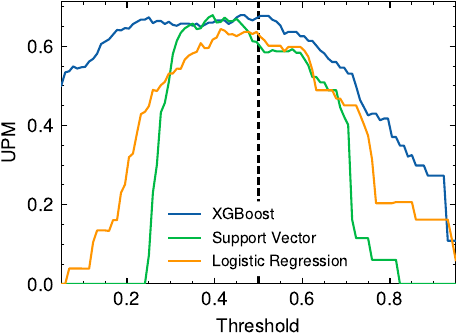}
    \caption{UPM is only slightly affected by the chosen decision threshold. UPM is calculated with different thresholds above which a prediction is rated positive.}
    \label{fig:upmthreshold}
\end{figure}

\begin{figure}
    \centering
    \includegraphics[scale=\imgscale]{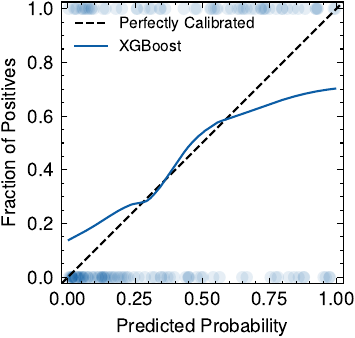}
    \caption{Calibration display for XGBoost based on the test data. The true class is shown in the scatter plot. This scatter plot is smoothed using LOWESS.}
    \label{fig:calibration}
\end{figure}

In terms of \gls{upm}, XGBoost outperforms the other classifiers with statistical significance. \cref{fig:modelcomparison} shows that XGBoost is generally stochastically dominant over \glspl{svc} and logistic regression ($\alpha=0.05$).
XGBoost UPM was persistently superior and robust across the threshold range (\cref{fig:upmthreshold}).
To provide a concrete example, \cref{fig:xgboost_cm} shows the confusion matrix for XGBoost with a threshold of \num[round-mode=figures,round-precision=2]{\xgbcThresAtMinSeventySens}. This threshold was chosen to obtain at least \SI{70}{\percent} specificity.
XGBoost was well-calibrated in the sense that a higher predicted probability generally corresponded to a higher risk of failed non-union healing, (\cref{fig:calibration}). However, XGBoost does not exactly estimate the probability of failed healing. The prediction has a calibration odds ratio of \rnum{\calibrationoddsratio} vs. the actual risk of failed non-union healing. Due to this imperfect calibration, the XGBoost confidence cannot be used to estimate the true healing likelihood (i.e., aleatoric uncertainty).

\begin{figure}
    \centering
    \includegraphics[scale=\imgscale]{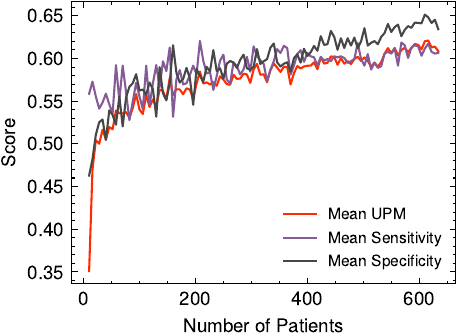}
    \caption{UPM, sensitivity, and specificity of XGBoost increase with more patients in the training dataset. Each data point is generated by sampling the entire training dataset 25 times, training the model, and calculating metrics on the test dataset. The threshold is chosen at 0.26.}
    \label{fig:samplecount}
\end{figure}

Finally, \cref{fig:samplecount} shows that artificially decreasing the number of patients in the training dataset only slightly decreases the performance, as measured by~\gls{upm}. This further underlines that our XGBoost model performs well with real-world clinical datasets which are limited in size.

\section{Conclusion}
\Gls{ml} models, especially XGBoost, are suited to identifying patients at risk of failed healing after non-union revision surgery.
Clinically relevant predictive performance has been demonstrated using data from a single center. 
Future research should consider investigating the performance with larger, multi-center datasets.
Our results pave the way to identifying patients at risk of failed non-union healing and potentially allow for more personalized treatment of the debilitating non-unions encountered in trauma surgery.

\section*{Acknowledgment}
We thank Dr. Dr. Matthias Reumann for his feedback throughout the research process.

\bibliography{references-zoterobetterbibtex}{}

\end{document}